\documentclass[11pt]{article}

\usepackage[preprint]{acl}

\usepackage{times}
\usepackage{latexsym}

\usepackage[T1]{fontenc}

\usepackage[utf8]{inputenc}

\usepackage{microtype}

\usepackage{inconsolata}
\usepackage{listings}

\usepackage{graphicx}
\usepackage{array}

\usepackage{adjustbox}
\usepackage{multirow}
\usepackage{paralist}
\usepackage{booktabs}

\title{Understanding the Impact of Linguistic Realization Choices \\ on LLM Stance with Causal Tracing}

\author{Langchen Huang \\ \And
   \textbf{Sebastian Pad\'o} \\ 
   Institute for Natural Language Processing, University of Stuttgart \\
   \texttt{\{langchen.huang|pado|franziska.weeber\}@ims.uni-stuttgart.de} \\ \And
   Franziska Weeber
   }
   
\begin{document}
\maketitle
\begin{abstract}

Large language models (LLMs) are known to be sensitive to prompt and input formulations. However, existing studies have focused on lexical realization and largely ignored constructional choice. This paper studies whether linguistic construction can systematically shift LLM decisions and where these shifts can be causally localized inside the model. We use political stance judgment as a meaning-sensitive case study and extend an English political statements dataset, resulting in six controlled linguistic rewrite types that preserve or invert the meaning of a statement. Experiments on four open-weight models show that stance instability affect both meaning-preserving and meaning-inversing rewrites. Because output shifts reveal that rewrites affect stance, but not where in the model, we apply activation patching, where activations from the original statement are substituted into the forward pass for the rewritten statement and measure which components recover the original stance distribution. The results show that mid-to-late decoder layers, especially block outputs at the final prompt position, provide the strongest restoration signal.

\end{abstract}

\section{Introduction}
\label{sec:intro}

A growing body of work has shown that LLM behavior can be brittle: Small surface-level perturbations to an input can yield disproportionately different outputs \citep{ceron-etal-2024-beyond,bang-etal-2024-measuring,rupprecht2025prompt}.This brittleness is particularly striking from a linguistics perspective. Natural language allows the same proposition to be expressed through many alternative realizations, often without changing the underlying semantics. However, previous work treats reformulation broadly, often through paraphrases or prompt wording changes, rather than isolating specific linguistic structures that systematically preserve or reverse sentence meaning. Therefore, a desirable property for downstream tasks that depend on meaning (e.g., stance detection or policy agreement) is semantic invariance: If the intended semantics of a statement are preserved, the model's decision should remain stable. This paper studies six linguistic rewrites with a focus on structural variation: Two meaning-reversing edits, namely explicit negations and semantic opposites, and four meaning-preserving rewrites, namely active/passive conversions, it-clefts, wh-clefts, and support verb constructions (SVC).

As downstream task, we focus on political stance judgment, where the model indicates the extent to which it agrees or disagrees with a policy statement. We use this task because it provides an interpretable decision signal for testing whether controlled linguistic rewrites affect model judgments in a meaning-sensitive setting.

Beyond establishing whether such response shifts result from structural variation caused by rewrites, we ask where they arise inside the model. Prior studies show that outputs vary under reformulation, but they do not identify the internal components that causally mediate these changes. The goal of this paper is to move from black-box observations of stance instability toward a more mechanistic account~\citep{meng2022locating,zhang2024towards,poonia-jain-2025-dissecting}. With this mechanistic account, it would be possible to attribute failures to specific internal components. 
 Concretely, the paper intends to address two research questions:
 \begin{itemize}
  \item \textbf{RQ1:} Which controlled linguistic rewrites could trigger a change in political opinion expressed by an LLM?
  \item \textbf{RQ2:} Where inside LLMs, at which layers and submodules, do these linguistic rewrites causally determine the response shift?
 \end{itemize}

This paper makes the following contributions:
 \begin{itemize}
  \item \textbf{Dataset extension with controlled linguistic rewrites for political stance evaluation}. This paper releases an extension of the English version of ProbVAA \citep{ceron-etal-2024-beyond} that adds four non-semantic, linguistically motivated meaning-preserving variants per base statement, alongside the two existing meaning-inverting variants in ProbVAA. 
  \item \textbf{Identification of linguistic rewrites that can trigger a political stance change}. This paper analyzes how various linguistic rewrites shift political stance in LLMs. The analysis reports reliability-oriented metrics (flip rates and Wasserstein distance) and uses variance decomposition to identify the source of stance variation.
  \item \textbf{Intervention-based analysis of sufficient restoration sites.} Using activation patching, this paper reports experimental results on the internal components that might cause stance flips.\footnote{The dataset and related code are available here: \url{https://github.com/nobody294/linguistic_rule_triggers}}
 \end{itemize}

\section{Related Work}
\label{sec:related work}
\paragraph{Sensitivity of LLM responses to input properties.} Prior evaluations show that LLMs shift their response when prompts are paraphrased, minimally edited, or reformulated.~\citet{ceron-etal-2024-beyond} introduce ProbVAA, a corpus of voting advice application statements with semantics-preserving paraphrases as well as semantics-reversing edits (explicit negation and semantic opposites). They find larger models to be more robust than smaller ones, but still susceptible to wording changes, specifically meaning-reversing ones.~\citet{haller2024yes} further demonstrate substantial paraphrase instability on Political Compass Test (PCT) statements, finding that some paraphrased statements induce large swings and even flip the binary decision. ~\citet{rottger-etal-2024-political} show that values and opinions evaluations based on the PCT are also sensitive to how models are instructed to adhere to a closed-ended multiple-choice response format. 

Beyond political questionnaires, similar sensitivity appears in normative survey settings:~\citet{rupprecht2025prompt} examine brittleness using closed-ended, value-oriented survey questions from the World Values Survey (WVS). Their results show that although larger models are generally more robust, all models remain sensitive to semantic variations like paraphrasing, to the order of answer options and to combined perturbation (paraphrased question + reversed-order answers). To our knowledge, no previous work systematically tested linguistic meaning-preserving rewrites instead of random paraphrases.

\paragraph{Mechanistic interpretability of stance shifts in LLMs.} Previous findings on the lack of robustness to LLM paraphrases remain largely behavioral. To go beyond behavioral evaluations, our work draws on interpretability methods.~\citet{zhang2024towards} systematically evaluate activation patching, a causal intervention that replaces selected hidden states in a corrupted run with those from a clean run. They evaluate several non-political controlled tasks. Their results show that activation patching can recover meaningful localization signals on these tasks.~\citet{meng2022locating} successfully use activation patching to locate decisive MLP layers for a subject-driven factual association task.~\citet{amirzadeh-etal-2024-language} apply a value-based variant of the method to gender-pronoun disambiguation with multiple contextual gender cues. Their study uses patching-based interventions to test whether cue information encoded in the target word representation is actually used when the model predicts the target pronoun. Their findings suggest that activation patching can connect representational information flow with prediction-level effects on model outputs.~\citet{poonia-jain-2025-dissecting} apply activation patching in a multiple-choice question answering setting on MMLU to investigate how persona prompts alter reasoning on an objective task, and their results show that activation patching was effective for localizing persona-related computation.

\begin{table*}[t]
    \centering
    \setlength{\tabcolsep}{5pt}
 \small{   
    \begin{tabular}{llp{12cm}}
        \toprule
        \textbf{Rewrite} & \textbf{Group} &  \textbf{Statement Example} \\ \midrule
         (Original) & & Organizers of events should be able to request a vaccination certificate upon entry. \\
         Negation & Reversing &  Organizers of events should \textbf{not} be able to request a vaccination certificate upon entry. \\
         Opposite & Reversing & Organizers of events should be \textbf{unable} to request a vaccination certificate upon entry. \\ \midrule
         Passive & Preserving & Vaccination certificate should be able \textbf{to be requested} by organizers of events upon entry. \\
         It-clefts & Preserving &  \textbf{It is} the vaccination certificate that organizers of events should be able to request upon entry. \\
         Wh-clefts &  Preserving & \textbf{What} the organizers of events should be able to request is a vaccination certificate upon entry. \\
         SVC &  Preserving & Organizers of events should be able to \textbf{make a request} for a vaccination certificate upon entry. \\ \bottomrule
    \end{tabular}}
    \caption{Example statements for each type of linguistic rewrite (Group: Meaning-preserving or -reversing). Rewrites below the line are our new ones.}
    \label{tab-variants-examples}
    
\end{table*}

\section{Method and Experiments}
\label{sec:method-experiments}
\subsection{Models}
We experiment with four open source LLMs: Gemma-3-4B-IT, Gemma-3-12B-IT, Qwen3-4B and Qwen3-14B. They have shown to perform well for their size \citep{team2025gemma,yang2025qwen3}, the weights are publicly accessible, and they are small enough to make local inference feasible under the computational constraints of this study. The models also span a scale from 4B-14B to consider the relationship between robustness and model size.

\subsection{Data}
\label{sec:data}
Our data is an extension of the ProbVAA dataset \citep{ceron-etal-2024-beyond}. ProbVAA contains 239 policy statements from voting advice applications as well as meaning-reversing rewrites and random paraphrases. Being designed for evaluating the reliability and consistency of LLM stances on political statements directly matches our focus on political stance judgment. Also, its items are single sentence statements, making them appropriate units for controlled linguistic rewrites since the same policy content can be reformulated while keeping the target judgment comparable. 

We select controlled linguistic rewrites that are possible to change model decisions under minimal surface edits. A candidate rewrite is selected if it: (a) has documented empirical impact on model behavior in prior works; (b) is a minimal edit that targets a specific capability (e.g., polarity, argument realization); (c) could be realized across topics, enabling broad coverage; and (d) introduces no semantic change or systematically inverts it. Our chosen linguistic rewrites set is presented in Table~\ref{tab-variants-examples}, divided into a polarity reversing and a polarity preserving group. Explicit negations tests whether models track syntactically marked logical polarity changes~\citep{jang-lukasiewicz-2023-consistency}, whereas semantic opposites test whether models track polarity reversals expressed through lexical choice~\citep{ceron-etal-2024-beyond}. Active/passive conversions changes argument order and surface prominence while preserving event roles, so shifts under this rewrite would suggest sensitivity to linear realization rather than only to predicate-argument structure~\citep{kim-etal-2018-teaching}. It-clefts and wh-clefts manipulate information structure, allowing us to test whether focus packaging alone can affect stance judgments~\citep{wang2024rupbench,li-etal-2020-linguistically}. Support/light verb constructions (SVC) paraphrase a verbal predicate as a light/support verb plus a deverbal noun, testing whether models remain stable under multiword predicate realizations~\citep{cap-etal-2015-account}.

\begin{table}[tb!]
  \centering
  \begin{tabular}{cc}
    \toprule
    \textbf{Category} & \textbf{Count} \\ \midrule
    original & 239 \\
    explicit negations & 239 \\
    semantic opposites & 239 \\
    active/passive & 194 \\
    it-clefts & 236 \\
    wh-clefts & 238 \\
    SVC & 122 \\ \bottomrule
  \end{tabular}
  \caption{Number of statements per linguistic rewrite.}
  \label{tab-1}
\end{table}

In the experiments, we use the English version of ProbVAA and their corresponding explicit negation variants and semantic opposite variants. We use Qwen3-8B~\citep{yang2025qwen3} to generate the other four variants via few-shot prompting (see Appendix~\ref{sec:generation-prompt}). We manually validate all generated statements. Table~\ref{tab-1} reports the number of statements in each variant where we could generate a valid transformation. The main outliers are passives and support verb constructions, which are not available for all sentences.

\subsection{Output Shape and Prompt Design}
\label{sec:output-representation}
The target behavior in this work is political stance elicitation where a user or model needs to judge how strongly they agree or disagree with a policy statement, so there is no ground truth response. Since we want to capture nuanced differences in the judgment, we use a 7-point Likert scale response format, which preserves a midpoint and provides three graded intensities on each side. This scale is common and methodologically accepted in similar studies such as~\citet{jeoung2025examining}. The intervention pipeline elicits stance using a constrained 1-7 Likert prompt (see the prompt in Appendix~\ref{sec:responses-prompt}). This turns the response readout into the model's next-token distribution at a well-defined answer position: the final input token index. This is the position whose hidden state produces the logits for the first generated token, which in this setup is intended to be a Likert digit. Operationally, restricting the output to a single-token set improves reproducibility and comparability. This setting is also consistent with classic mechanistic interpretability protocols that evaluate interventions via changes in the output logits/probabilities of specific candidate answer tokens \citep{zhang2024towards}.

The set of digit tokens is $D = \{1, 2, \ldots, 7\}$ with the standard interpretation: 1: strongly disagree, 2: disagree, 3: slightly disagree, 4: neutral, correspondingly 5--7: slightly/(no modifier)/strongly agree. Let $z(x) \in R^{|V|}$ be the next-token logits at answer position $p$. A 7-dimensional logit vector is formed by slicing out the digit logits:
\begin{equation}
  z_D(x) = (z_{t(1)}(x), z_{t(2)}(x), \ldots, z_{t(7)}(x))
\end{equation}
We then apply Softmax to the digit logits and use the resulting probability vector as the output representation of the model.

\subsection{Experiment 1: Responses Generation and Stance Judging}
\label{sec:responses-generation}
We prompt each model with each base statement and each valid linguistic rewrite. The prompt follows the format in Section~\ref{sec:output-representation}. Based on prior work on reliable LLM evaluations \citep{ceron-etal-2024-beyond,zhong2025evaluating}, we sample 30 stochastic outputs per prompt (temperature = 0.8) and treat them as Monte Carlo draws from the model's conditional response distribution. Each output is a 7-point Likert digit $d$. We then aggregate the 30 responses into a centered stance score $s \in [-1, 1]$ using the mapping $s = (d - 4) / 3$. To quantify uncertainty, we perform a nonparametric bootstrap with 1,000 resamples and obtain a 95\% confidence interval (CI). These intervals were passed to the clear-leaning rate and flip rate evaluations defined in Section~\ref{subsec:evaluation}.

\subsection{Experiment 2: Activation Patching}
For a given base statement and its variant, two model behaviors are observed: a clean response to the base statement, and a corrupt response to the variant. Activation patching tests whether swapping a specific internal activation from a clean run into a corrupt run is sufficient to restore the clean behavior at the output. It requires caching the internal activations of the clean forward pass that will later be substituted into the corrupt run. In the experiments, we cache three activation sites that are commonly investigated in transformers for each decoder layer~\citep{meng2022locating}: decoder layer output (block output), attention sublayer output, and Multilayer Perceptron (MLP) sublayer output.

\begin{figure*}[t]
  \centering
  \includegraphics[width=0.9\linewidth]{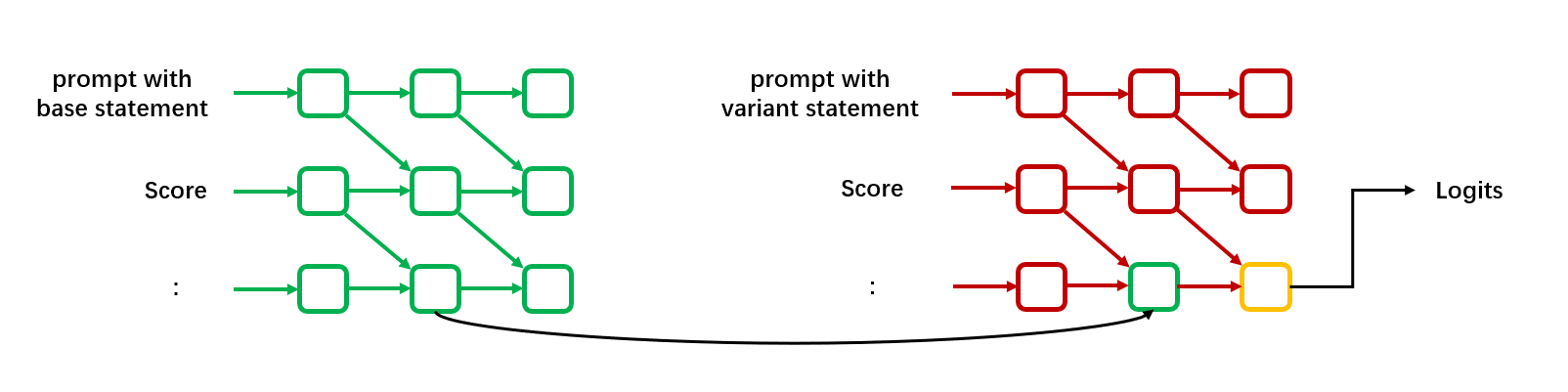}
  \caption{The workflow of activation patching, running the intervention procedure on every defined component. Each square represents the activation of a token in one component at one layer. ``Score:'' is also part of the prompt.}
  \label{fig:activation patching}
\end{figure*}

The base and variant prompts can differ in tokenization length (e.g., explicit negation rewrites introduce extra tokens). To avoid introducing additional alignment assumptions, the experiments therefore intervene only at the final prompt position preceding the generated answer token~\citep{zhang2024towards}. Figure~\ref{fig:activation patching} visualizes the workflow.

\subsection{Evaluation}
\label{subsec:evaluation}
\paragraph{Clear-leaning rate and flip rate.} Following reliability-aware evaluation in prior work on political stance stability in LLMs~\citep{ceron-etal-2024-beyond}, the experiment defines a practically negligible region around neutrality, which is $[-0.1, 0.1]$ for a $[-1, 1]$ value range. We consider a prompt to elicit a clear leaning only if the 95\% CI lies entirely outside the interval $[-0.1, 0.1]$. Otherwise, the evidence is insufficient to claim a clear leaning.

To quantify directional stance instability under controlled linguistic rewrites, we calculate the flip rate, defined as the proportion of base-variant pairs whose binarized stance polarity violates the expected polarity relation. If the $95$\% CI lies in $[0.1, 1]$, the model agrees with the statement (positive polarity); if the $95$\% CI lies in $[-1, -0.1]$, the model disagrees with the statement (negative polarity). We compute the flip rate only over eligible base-variant pairs for which both the base statement and the rewritten variant receive a clear positive or negative polarity. Pairs in which either side is unclear are excluded from the flip rate calculations.

\paragraph{Variance Decomposition.} To better understand what causes variation in LLM responses, this paper complements the evaluation with a variance decomposition of stance scores. The procedure follows the diagnostic framework proposed by~\citet{kunievsky2026measuring}. Let $s$ denotes base statements, $v$ denotes linguistic rewrite variants, $r$ denotes repeated samples, and $E$ denotes the expectation. For each $(s, v, r)$ the model generates a 1-7 Likert score $y_{s,v,r}$, define $\bar{y}_{s,v} = E_r[y_{s,v,r}]$ and $\bar{y}_s = E_v[y_{s,v}]$. Variability is decomposed into three components:
\begin{compactitem}
  \item \textbf{Purpose Sensitivity (PS)} captures between-item variation, i.e., how much the model's average stance differs across policy statements. PS quantifies the extent to which the model's stance is meaningfully conditioned on the substance of the policy item (topic/issue content).
  \item \textbf{Articulation Sensitivity (AS)} captures within-item variation due to linguistic rewrites, i.e., how much the model's mean stance changes when the same underlying statement is expressed with different syntax/information structure. AS is the primary diagnostic for stance shifts under linguistic rewrites: Higher AS indicates greater vulnerability to prompt phrasing.
  \item \textbf{Model Uncertainty (MU)} captures within-prompt variation across repeated sampling, i.e., the noise induced by stochastic generation. MU could help distinguish linguistic rewrites effects from mere sampling jitter.
\end{compactitem}
Detailed variance decomposition formulas are in Appendix~\ref{sec:VD-formulas}. For pairwise differences, the variance of a difference between two quantities from the same component is twice the component variance. Therefore, Likert point shifts are reported as $\sqrt{2PS}$, $\sqrt{2AS}$, and $\sqrt{2MU}$.

\paragraph{Wasserstein Distance ($W_1$).} The patching experiment intervenes not on a single correct label, but the entire model-implied distribution over the 7-point Likert scale. Therefore, we evaluate using the 1-Wasserstein distance (also known as Earth Mover's Distance, EMD) since it explicitly uses the ground distance between adjacent ordered classes, penalizing probability mass shifts proportionally to how far they move along the ordinal axis ~\citep{nakov-etal-2016-semeval}.

\paragraph{Normalized restoration score.} Patching requires comparing intervention effects across many prompts and components. However, the gap between the base prompt and its variant can vary substantially across examples. Reporting only the raw distance reduction can therefore make cross-example comparisons misleading. To address this, we use a normalized restoration score that measures the fraction of the base-variant deviation removed by an intervention (see formulas in Appendix~\ref{sec:w1-restoration-formulas}). If the score $R \approx 1$, the intervention nearly restores the clean distribution. If $R \approx 0$, the intervention never restores the clean distribution. If $R < 0$, the intervention causes a larger difference.

\begin{figure*}[t]
    \centering
    \includegraphics[width=0.48\linewidth]{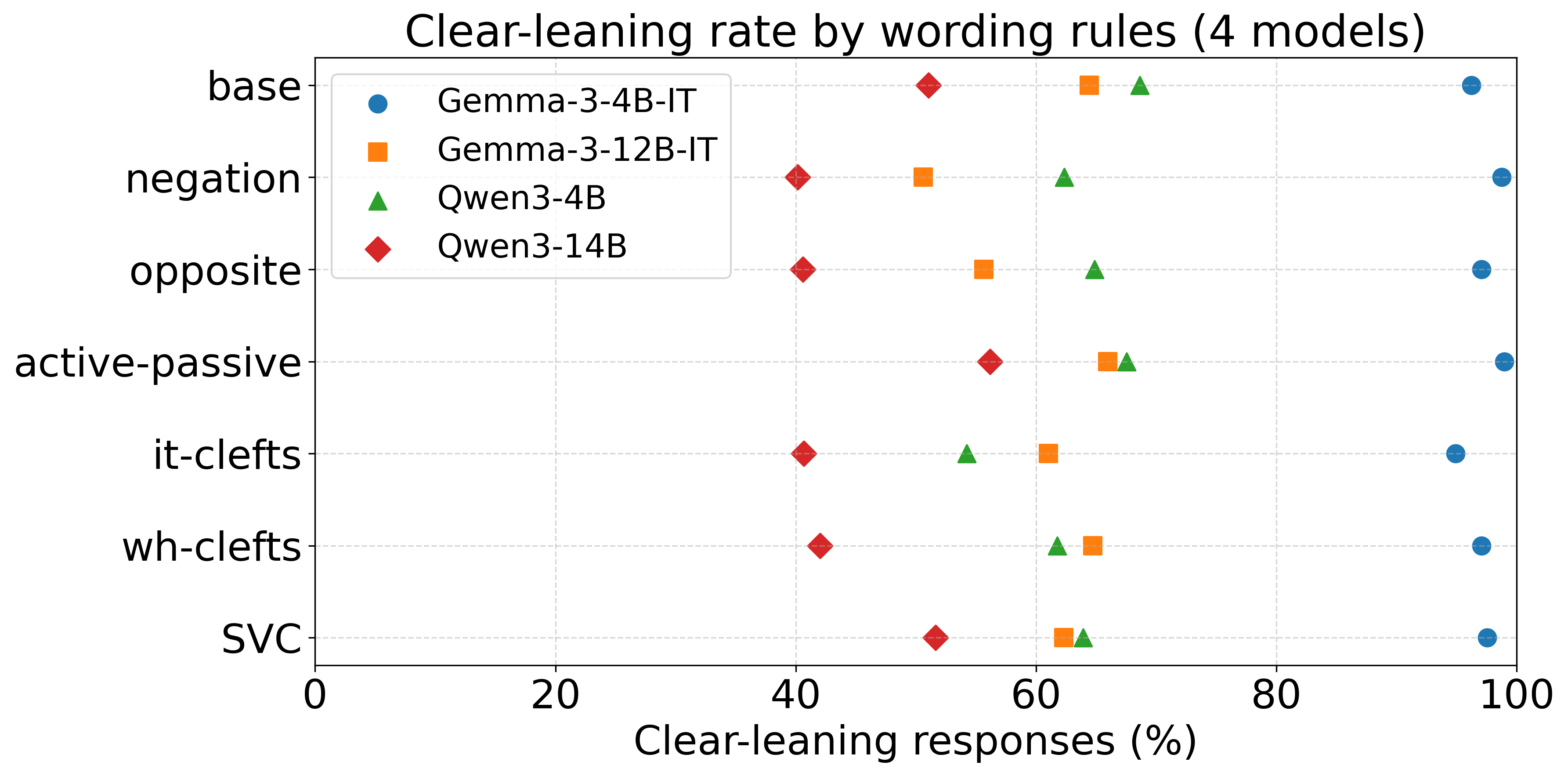}
    \includegraphics[width=0.48\linewidth]{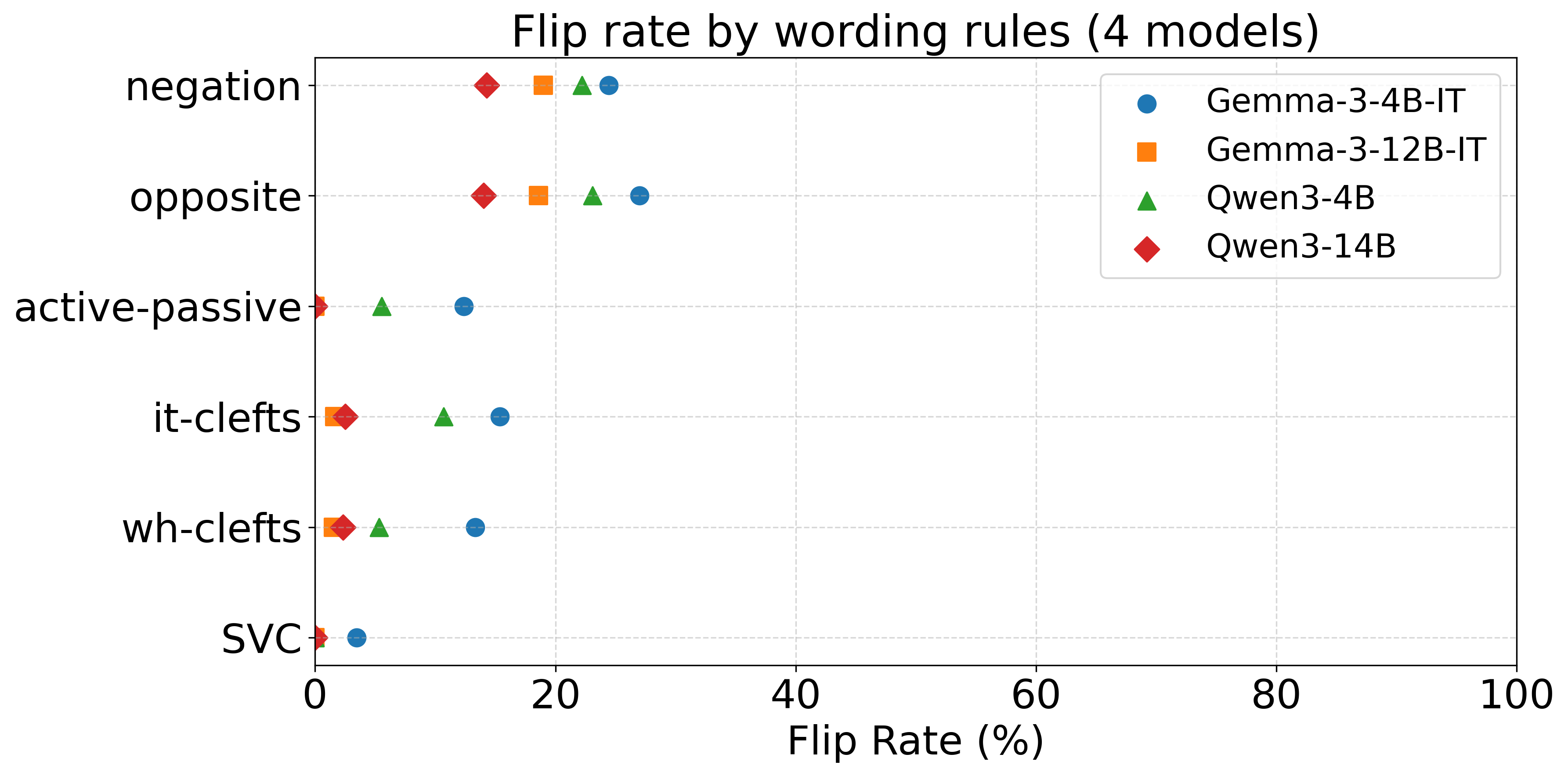}
    \caption{Clear-leaning rate and flip rate results.}
    \label{fig:results-clear-flip}
\end{figure*}

\section{Results and Discussion}

\subsection{Experiment 1: Response Generation and Stance Judging}
\label{subsec:results-stance judging}
\paragraph{Clear-leaning rate and flip rate.} The percentages of clear-leaning responses of the four models are shown in Figure~\ref{fig:results-clear-flip}. Gemma-3-4B-IT's clear-leaning response rate is close to one, while Gemma-3-12B-IT shows a clear stance in slightly more than half of all cases. With 54.2\% to 68.2\% of clear-leaning responses rate for Qwen-4B and 40.2\% to 56.2\% for Qwen-14B, the differences within the model family are smaller, but as for Gemma, the smaller model has more clear leaning responses.

The flip rates for each variant are also shown in Figure~\ref{fig:results-clear-flip}. The two model families present a similar trend again: The flip rate of Gemma-3-4B-IT ranges from 3.5\% to 27.0\%, while the flip rate of Gemma-3-12B-IT ranges from 0.0\% to 19.0\%. The flip rate of Qwen3-4B ranges from 0.0\% to 23.0\%, while the flip rate of Qwen3-14B ranges from 0.0\% to 14.3\%. The larger models produced a lower flip rate than the smaller models.

The clear leaning responses rates and flip rates indicate that the larger models more frequently yield near-neutral or more uncertain judgments and are less likely to be affected by linguistic rewrites. A plausible interpretation of this size effect is consistent with prior works on uncertainty estimation and calibration in LLMs: \citet{zhu-etal-2023-calibration} report that larger models are generally better calibrated and observe that, as parameter scale increases, the confidence distribution of model outputs tends to contract toward a smaller and lower range, interpreted as reduced overconfidence.

\paragraph{Variance decomposition.} The outcome of the variance decomposition in linguistic rewrites is shown in Table~\ref{tab-5}. As the results illustrate, sampling noise is small for the models. For Gemma-3-4B-IT, MU causes $\sqrt{2MU} \approx$ 0.24-0.42 Likert points shift, Gemma-3-12B-IT shows a similar range. For Qwen3-4B, MU causes 0.47-0.57 Likert points shift. For Qwen3-14B, MU causes 0.2-0.24 Likert points shift. For Gemma-3-4B-IT, AS of polarity preserving rewrites cause $\sqrt{2AS} \approx$ 0.63-1.23 Likert points shift. Their PS causes $\sqrt{2PS} \approx$ 2.10-2.31 Likert points shift. For Gemma-3-12B-IT, AS of the same rewrites induce 0.53-0.92 Likert points shift. The PS induced 1.61-1.75 Likert points shift. For the polarity reversing rewrites, the 4B model yield AS = 1.52 and 2.13, corresponding to 1.74 and 2.06 Likert points shift. The 12B model improves, with 1.58-1.64 Likert points shift. The same pattern was observed in the Qwen3 models.

\begin{table}[tb!]
  \centering
  {
  \small
  \begin{tabular}{ccccc}
    \toprule
    Linguistic Rewrites & Model & MU & AS & PS \\ \midrule

    \multirow{4}{*}{Original}
      & G-4B & 0.06 & -- & -- \\
      & 12B  & 0.04 & -- & -- \\
      & Q-4B & 0.12 & -- & -- \\
      & 14B  & 0.04 & -- & -- \\ \midrule

    \multirow{4}{*}{Negation}
      & G-4B & 0.03 & \textbf{1.52} & 1.48 \\
      & 12B  & 0.1  & \textbf{1.25} & 1.21 \\
      & Q-4B & 0.16 & 0.69 & \textbf{0.87} \\
      & 14B  & 0.06 & \textbf{0.57} & 0.53 \\ \midrule

    \multirow{4}{*}{Opposite}
      & G-4B & 0.06 & \textbf{2.13} & 1.70 \\
      & 12B  & 0.06 & \textbf{1.35} & 1.29 \\
      & Q-4B & 0.12 & 0.72 & \textbf{0.84} \\
      & 14B  & 0.06 & \textbf{0.53} & 0.51 \\ \midrule

    \multirow{4}{*}{Passive}
      & G-4B & 0.05 & 0.62 & \textbf{2.25} \\
      & 12B  & 0.05 & 0.19 & \textbf{1.36} \\
      & Q-4B & 0.11 & 0.23 & \textbf{0.96} \\
      & 14B  & 0.05 & 0.08 & \textbf{0.53} \\ \midrule

    \multirow{4}{*}{It-clefts}
      & G-4B & 0.09 & 0.76 & \textbf{2.29} \\
      & 12B  & 0.04 & 0.42 & \textbf{1.29} \\
      & Q-4B & 0.14 & 0.40 & \textbf{0.85} \\
      & 14B  & 0.05 & 0.18 & \textbf{0.49} \\ \midrule

    \multirow{4}{*}{Wh-clefts}
      & G-4B & 0.05 & 0.72 & \textbf{2.22} \\
      & 12B  & 0.04 & 0.34 & \textbf{1.36} \\
      & Q-4B & 0.13 & 0.35 & \textbf{0.82} \\
      & 14B  & 0.06 & 0.16 & \textbf{0.49} \\ \midrule

    \multirow{4}{*}{SVC}
      & G-4B & 0.06 & 0.2 & \textbf{2.67} \\
      & 12B  & 0.03 & 0.14 & \textbf{1.53} \\
      & Q-4B & 0.12 & 0.17 & \textbf{1.0} \\
      & 14B  & 0.04 & 0.04 & \textbf{0.53} \\ \bottomrule
  \end{tabular}
  }
  \caption{Variation decomposition results of the four models. G-4B and 12B stand for Gemma-3-4B-IT and Gemma-3-12B-IT, Q-4B and 14B for Qwen3-4B and Qwen3-14B. MU is model uncertainty, AS is articulation sensitivity, and PS is purpose sensitivity. AS (variant) and PS (variant) are AS (variant + original) and PS (variant + original). The highest value in each row is highlighted in bold. For the original statement, only MU is reported in the source table.}
  \label{tab-5}
\end{table}

The results of the variance decomposition show that the PS values and AS values are both greater than the MU values. This indicates that stance shifts cannot be explained by stochastic sampling. Also, issue content (between-item differences) remains the dominant driver of stance variability, but the articulation component is non-trivial: Syntax-level reformulations could shift the model by roughly half to over one Likert point on average. Also, the larger models show lower articulation sensitivity (smaller stance shifts under rewrites) and also lower between-item dispersion, consistent with a more centrist response result.

The results of experiment 1 demonstrate linguistic rewrite-specific effects. The models are almost invariant to SVC, which produce the lowest flip rates and AS values across model families, whereas polarity reversing rewrites produce higher flip rates and AS values than polarity preserving rewrites, showing that models do not fully capture the meaning of negations or opposite wording choices.

\subsection{Experiment 2: Activation Patching}
\label{subsec:results-patching}

\begin{figure*}[t]
  \centering
  \includegraphics[width=0.45\linewidth]{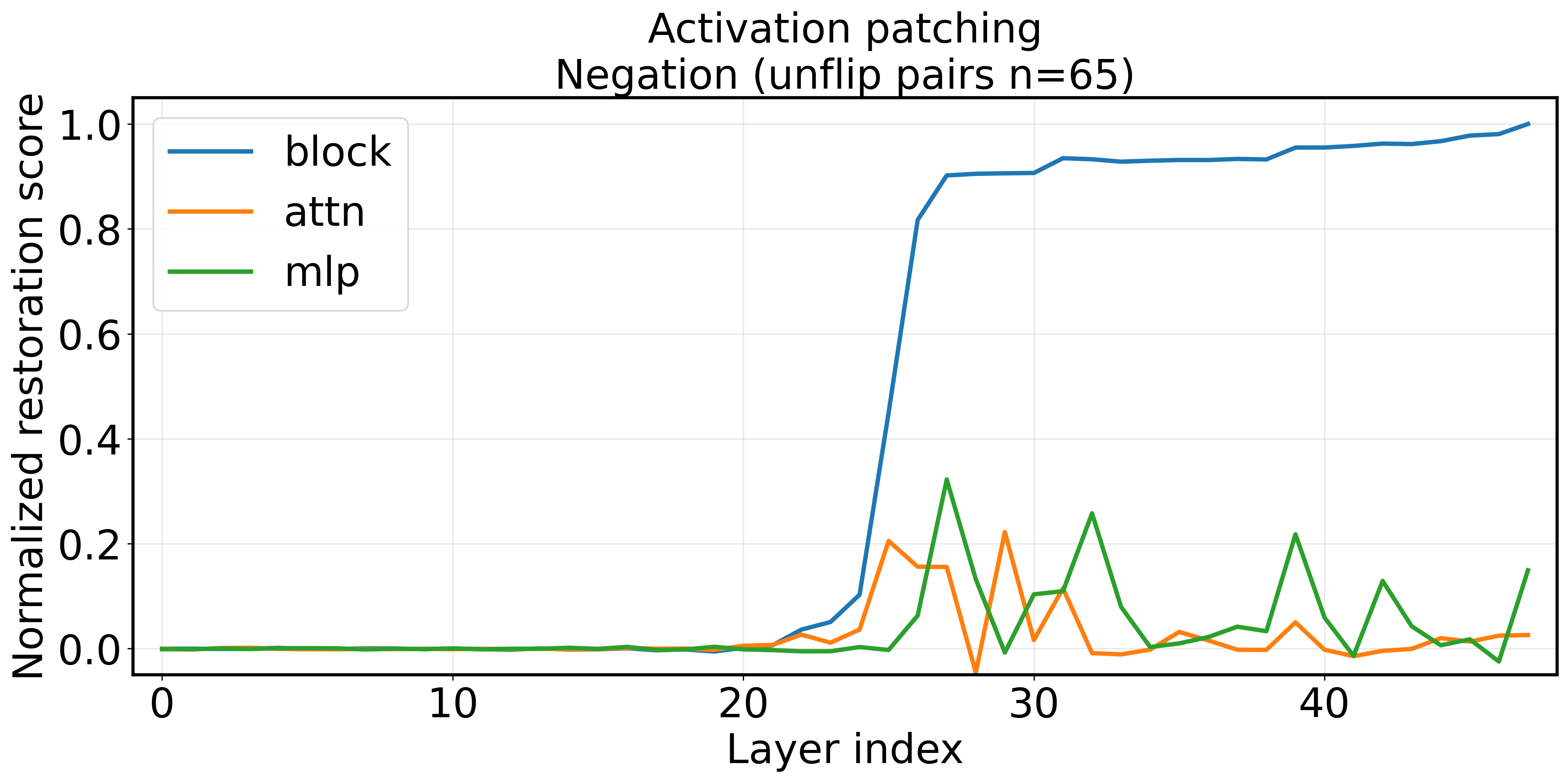}
  \includegraphics[width=0.45\linewidth]{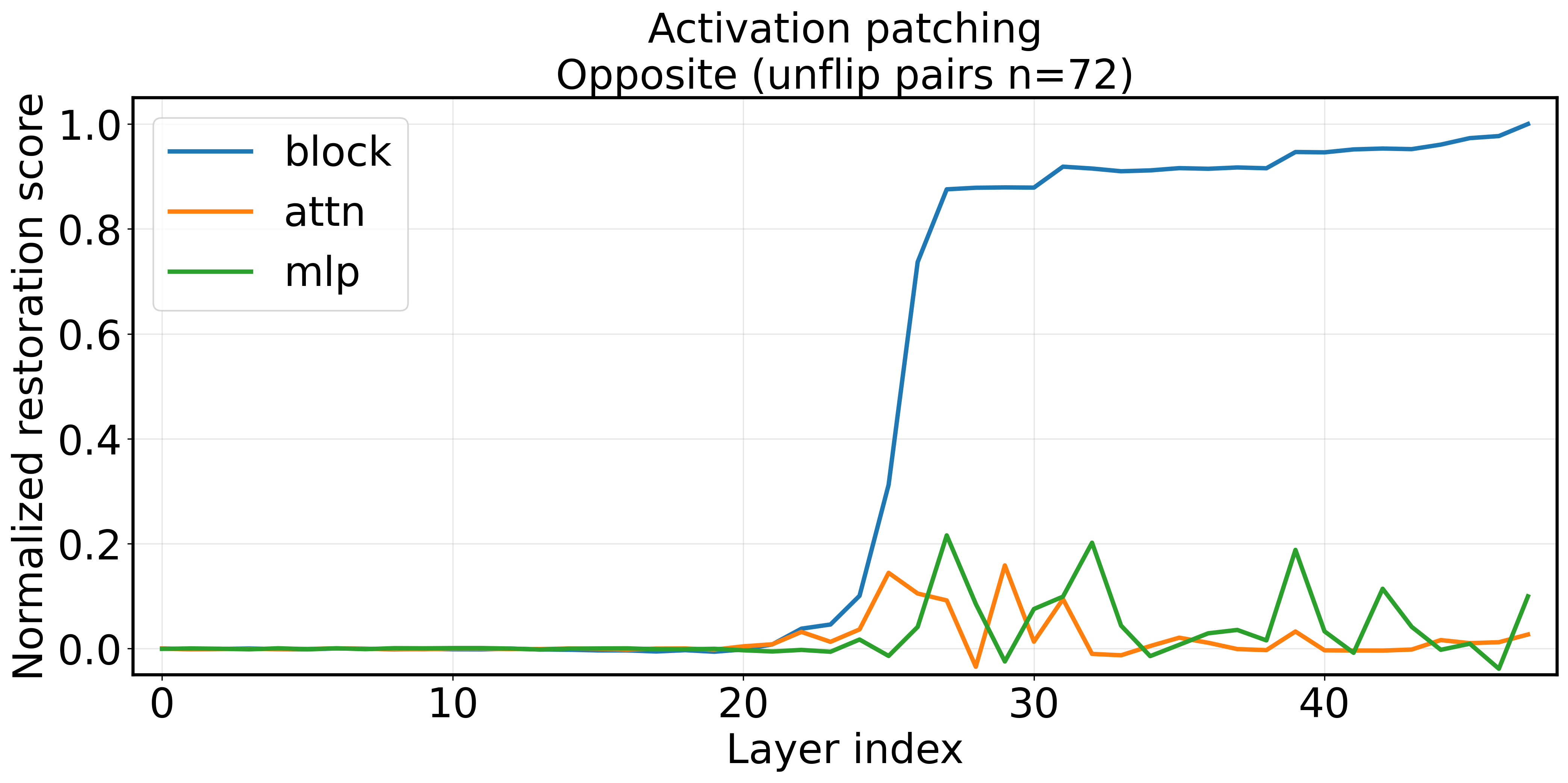}
  \includegraphics[width=0.45\linewidth]{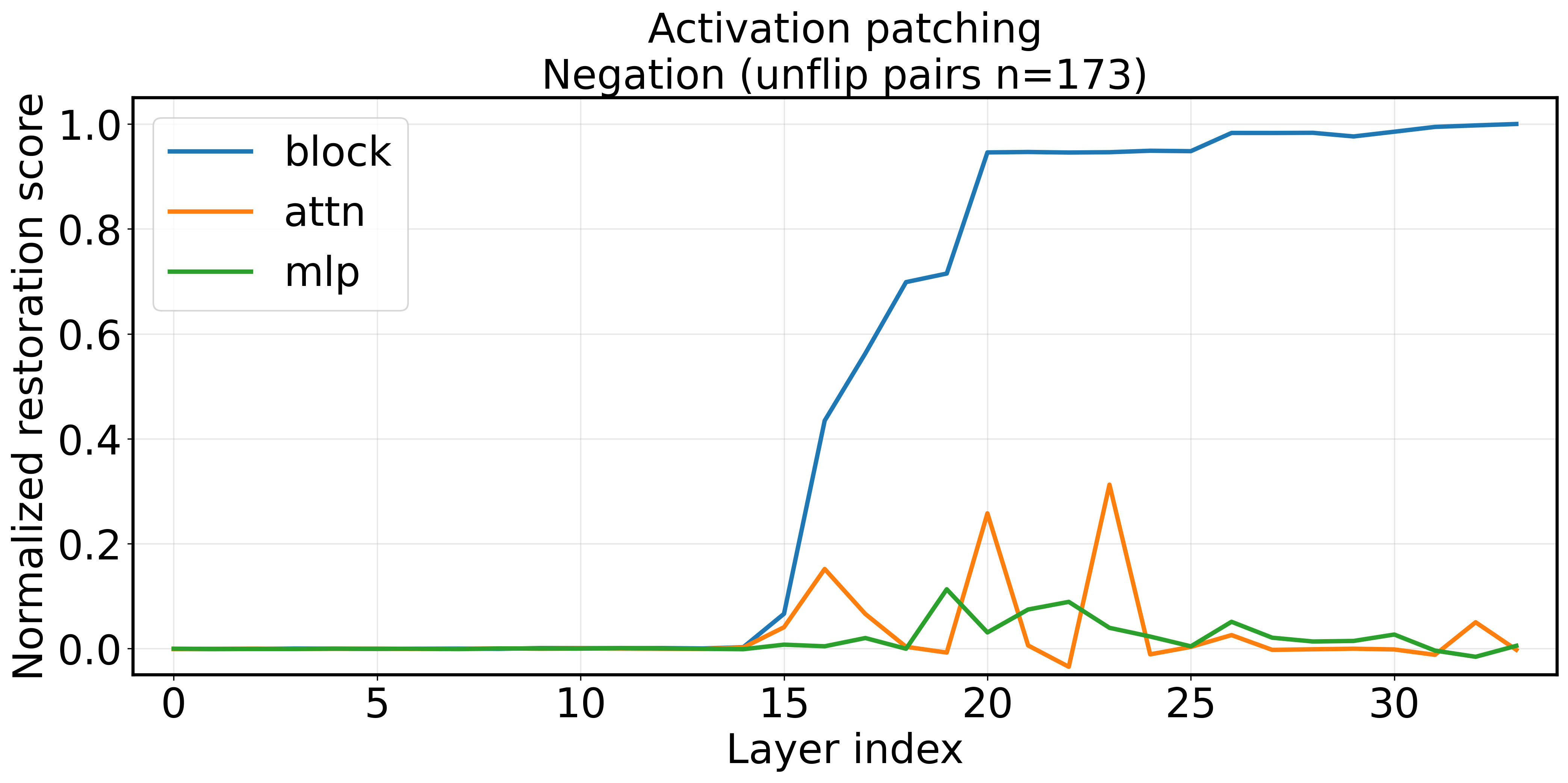}
  \includegraphics[width=0.45\linewidth]{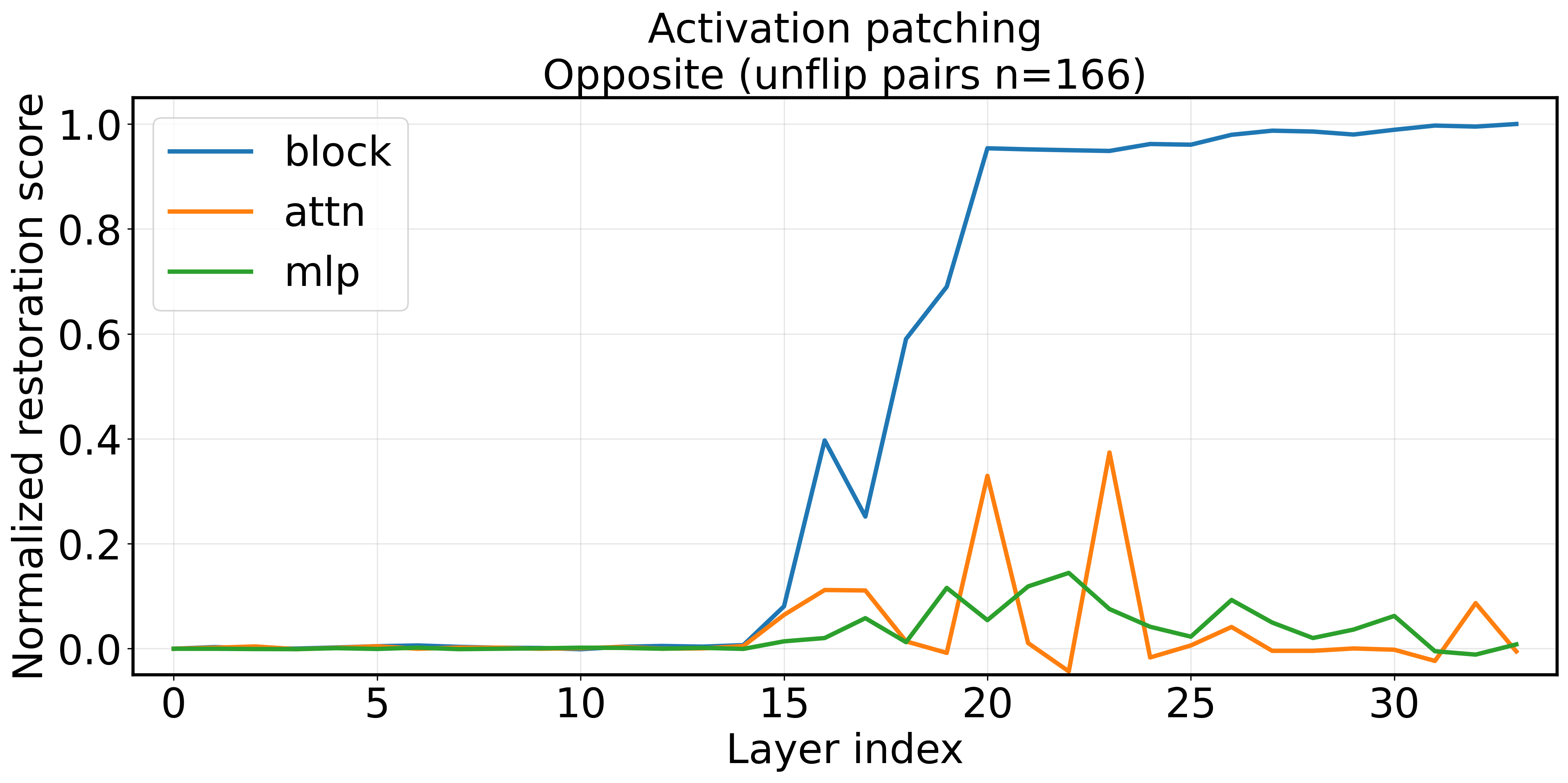}
  \includegraphics[width=0.45\linewidth]{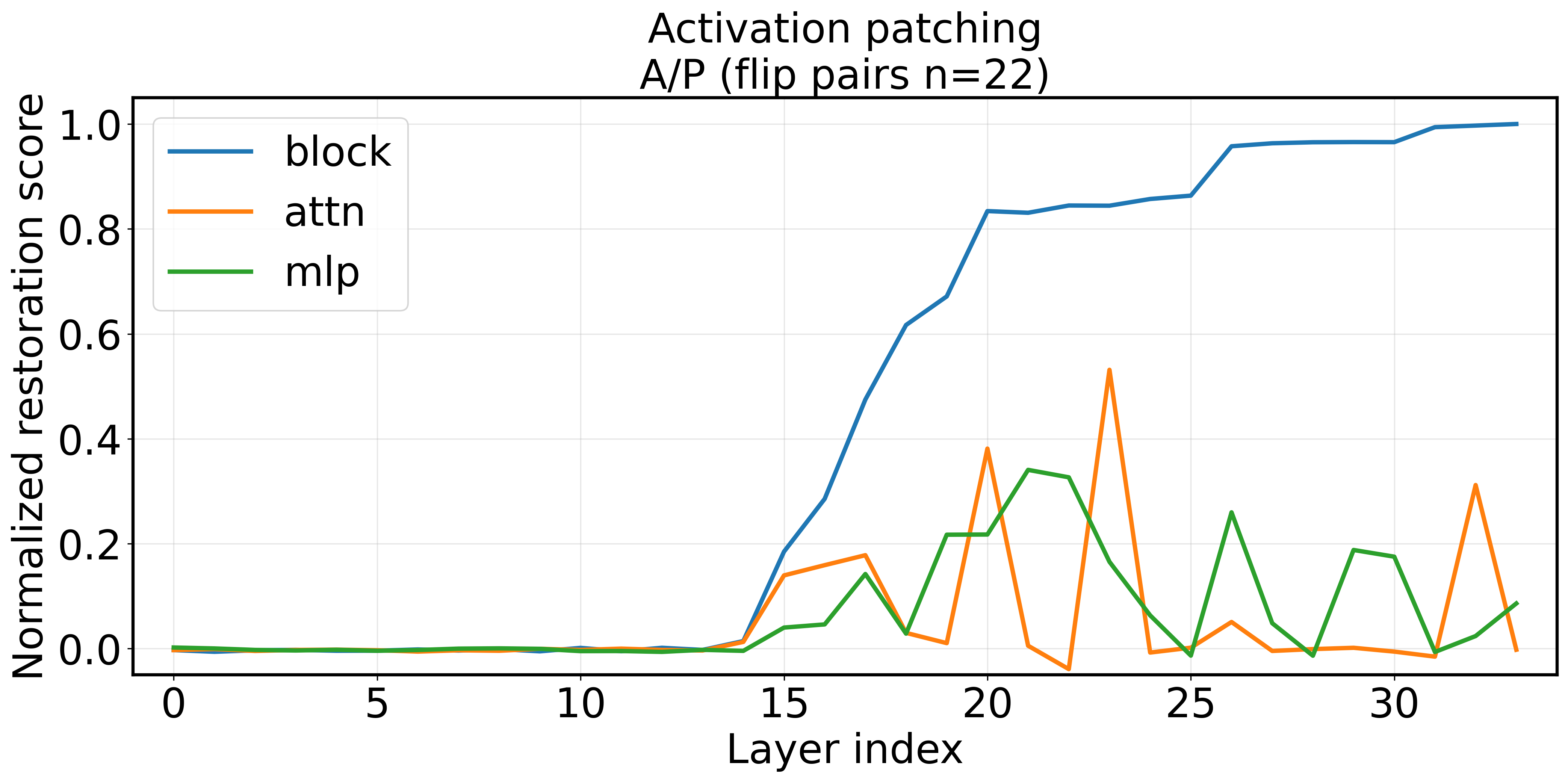}
  \includegraphics[width=0.45\linewidth]{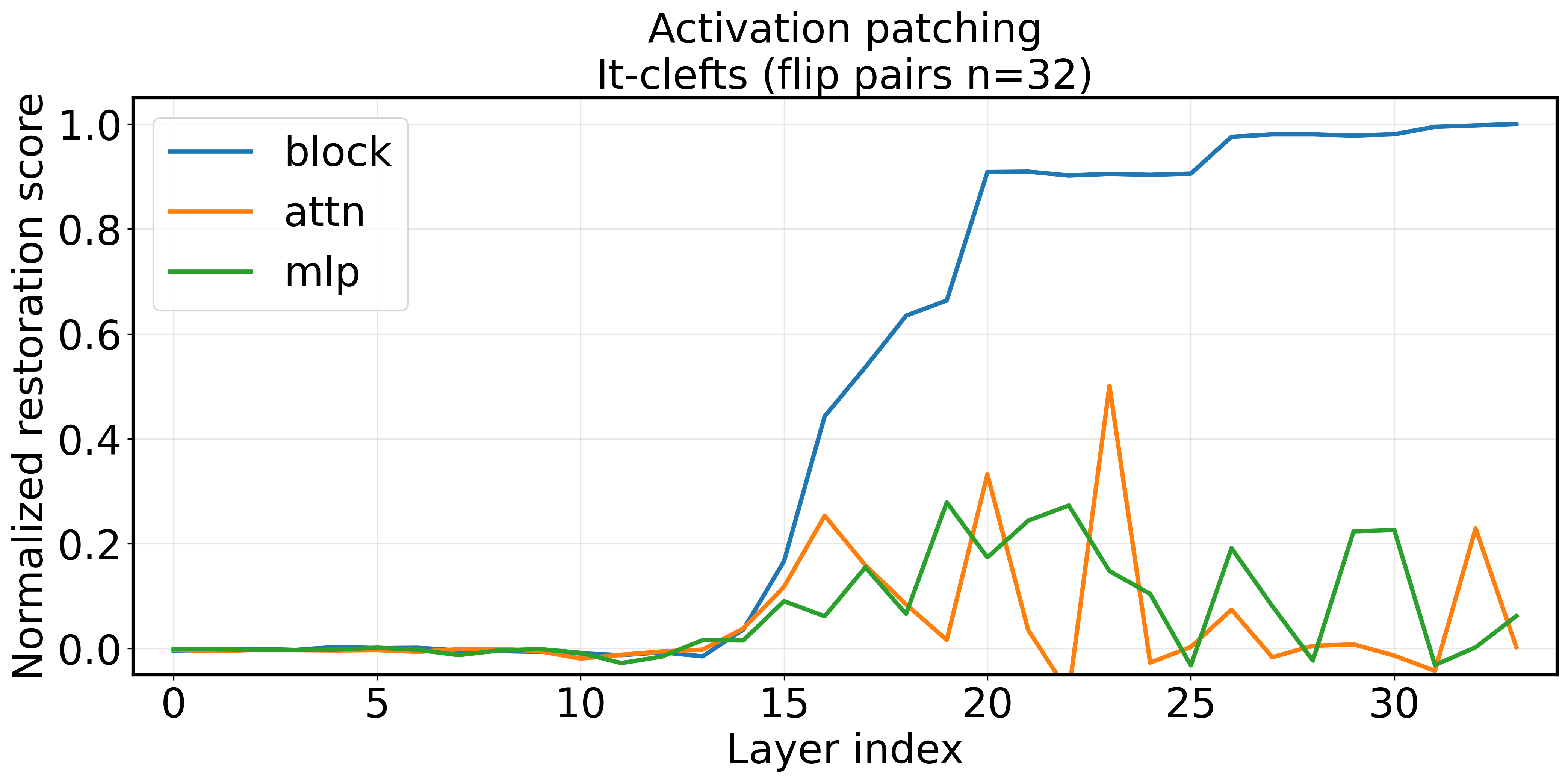}
  \includegraphics[width=0.45\linewidth]{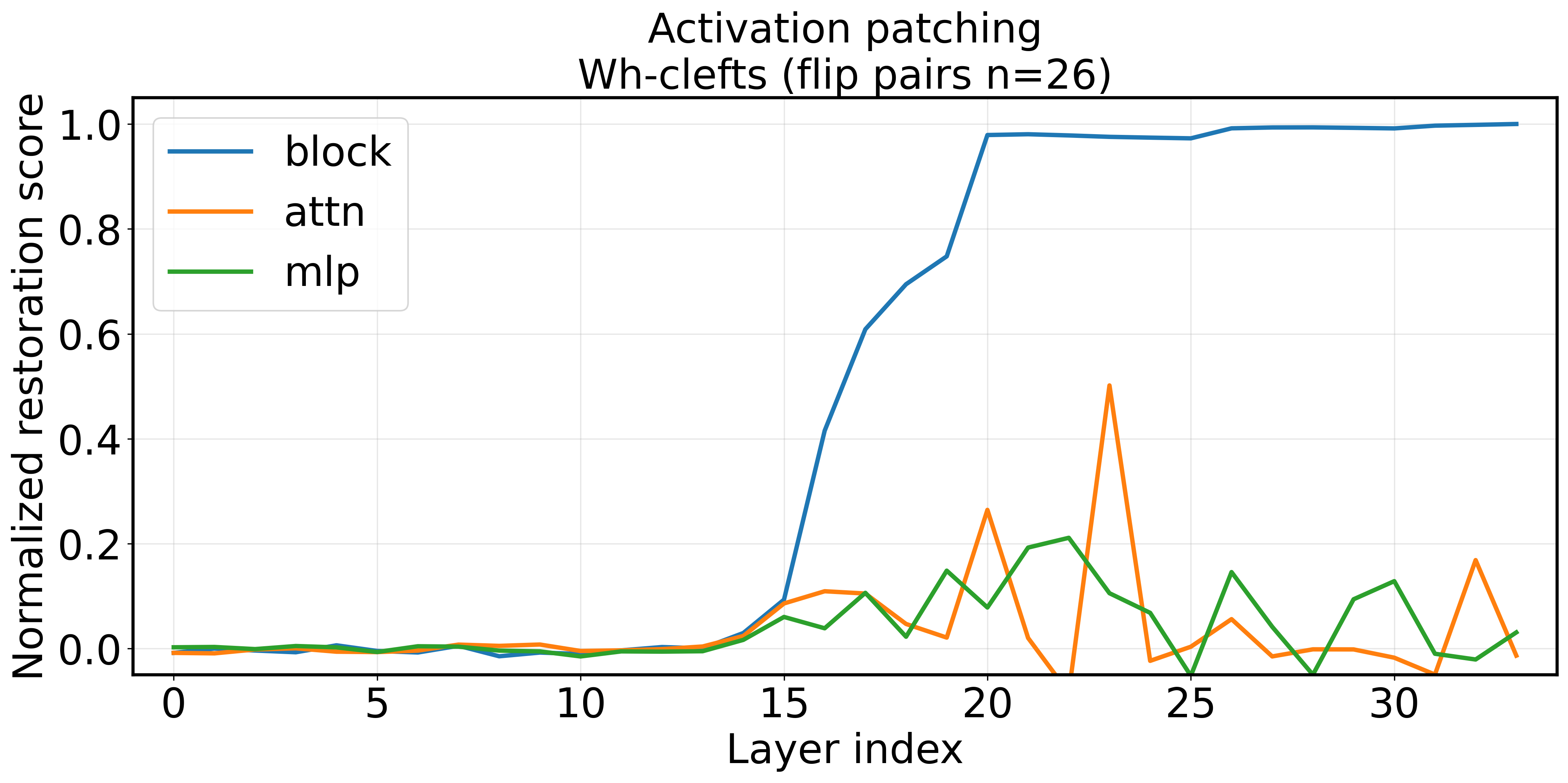}
  \caption{Activation patching results of Gemma-3-12B-IT (first row) and Gemma-3-4B-IT (second to fourth row). The normalized restoration scores are the mean value of each layer's component.}
  \label{fig:activation-patching-results-1}
\end{figure*}

Patching results are analyzed at the level of rewrite types. If a rewrite type produces only very few stance flips, the aggregate localization results would be too unstable to support a meaningful causal localization analysis. We therefore include only the following rewrite types for each model: For Gemma-3-4B-IT, we use all rewrite types except Support Verb Constructions (SVC). For Gemma-3-12B-IT and Qwen3-14B, we use the negation and opposite variants. For Qwen3-4B, we use negation, opposite, and it-clefts. The activation patching for the polarity reversing group was run on the pairs that exhibit the expected polarity reversal, because activation patching requires a measurable output difference.

Figure~\ref{fig:activation-patching-results-1} shows the patching results of Gemma-3-4B-IT and Gemma-3-12B-IT, while Appendix~\ref{sec:qwen3-patching} shows the results of the Qwen3 models, mirroring the pattern of the Gemma-3 models. For both model families, block-output patching produces the strongest and most stable restoration signal. This advantage is expected from the perspective of the transformer residual stream. In the standard residual-block formulation, each layer is a composition of a self-attention sublayer and a position-wise feed-forward (MLP) sublayer, with residual connections around sublayers \citep{vaswani2017attention}. Because the layer output is passed to all subsequent computation, it aggregates the additive contributions of both attention and MLP pathways. Consequently, patching the block output at the answer position overwrites a more complete summary of what the model has computed up to that point, whereas patching a single sublayer output replaces only one additive write into that state and allows the other pathway and any subsequent transformations to preserve or reintroduce the deviation.

For all models, patching attention or MLP activations of larger models mainly shifts responses back toward neutrality rather than restoring the clean stance. This is consistent with the behavioral results that the larger model produces fewer clear-leaning judgments and lower flip rates.

The figures show that rewrites-triggered political stance shifts in our tested models could be mediated by mid-to-late block outputs at the final prompt position. Similar outcomes have been found in emotion inference:~\citet{tak-etal-2025-mechanistic} show that patching hidden states can transfer the output emotion, with early-layer patching being ineffective, but with a clear rise in success after a critical layer and stable success around the mid-layers across models. 

Across polarity preserving and polarity reversing rewrites, the localization results are broadly similar. This suggests that although these rewrites differ linguistically and in whether they preserve or invert meaning, once they induce a change in political stance judgment, the downstream mechanism carrying that shift to the final score prediction is largely shared rather than rewrite-specific.

\section{Conclusion}
In this paper, we examined whether controlled linguistic rewrites alter political stance judgments in LLMs and whether activation patching can causally mediate such shifts. The results answer both research questions.

First, stance instability occurs for both meaning-reversing and meaning-preserving edits: Although explicit negations and semantic opposites are the strongest triggers, meaning-preserving rewrites also induce non-trivial changes in model judgments. This suggests that political stance evaluations are sensitive not only to propositional content, but also to linguistic realization. While prior work has mainly shown vulnerability to broad paraphrases, prompt wording, and elicitation format~\citep{ceron-etal-2024-beyond,haller2024yes,rottger-etal-2024-political}, our study isolates specific linguistic rewrites, showing that stance instability is not only a general paraphrase effect, but controlled constructional linguistic rewrites can also systematically shift stance judgments. The model-size contrast further suggests that robustness improves only partially with scale: larger models are less likely to produce flips, but they also more often move toward neutral or uncertain responses. Variance decomposition shows that these patterns are not primarily attributable to stochastic sampling, since substantive item differences remain the dominant source of variation and articulation effects are non-negligible.

Second, the paper contributes a mechanistic account of these response shifts. Activation patching shows that mid-to-late decoder block outputs at the final prompt position are often sufficient to restore the clean stance distribution, whereas attention and MLP interventions are substantially weaker. This extends prior behavioral work by demonstrating that rewrites-triggered stance shifts can be causally localized to specific internal components.

Overall, the findings show that linguistic rewrites can systematically alter LLM judgments in meaning-sensitive political evaluations. Future work should test whether the same pattern holds across additional model families, multilingual settings, and more fine-grained circuit analyses.

\section{Limitations} 
Our study faces several limitations. First, the empirical evaluation centers on four relatively small models from two model families, so the cross-family and size-effect evidence remains limited. Second, the findings are limited to one dataset with a closed-ended agreement rating task. Whether the results generalize to other datasets and tasks remains to be tested. Third, this dataset already includes the meaning-reversing rewrites, the lower robustness compared to our self-created meaning-preserving rewrites could also stem from the difference in data creation, but is unlikely due to our manual evaluation and strict rewriting rules. Fourth, small semantic changes might still occur for linguistic rewrites, even though we aimed to keep the semantics unchanged or inverted and manually evaluated the reworded statements. Fifth, we only evaluate English data, LLMs might react differently in other languages. Last, the mechanistic analysis localizes sufficient sites using activation patching at the final answer position, but it does not yet provide a full circuit-level account of how stance is computed across token positions and layers. Also, our activation patching analysis localizes sufficient restoration sites rather than the origin of the rewrite-sensitive computation.


\appendix

\section{Prompts}
\subsection{Variants Generation Prompt}
\label{sec:generation-prompt}

\begin{lstlisting}[caption={Prompt for active/passive conversions variants generation},breaklines=true,xleftmargin=0pt,basicstyle=\ttfamily\footnotesize]

 You are a controlled text rewriter. Your only job is to convert the base statement between active and passive voice. Generate in English only.

 Task: Convert the base statement between active and passive voice.

 Hard constraints:
 (1) Do exactly and only a voice transformation (Active<->Passive). Preserve all arguments, named entities, numbers, tense/aspect, modals, quantifiers, negation scope, and PPs.
 (2) If an agent exists, keep it (use a by-phrase in passive).
 (3) If voice transformation is inapplicable, set ``not applicable''=true.
 (4) Keep truth-conditional meaning intact. No paraphrasing beyond voice change.

 Output format (SINGLE JSON only, no extra text):

 {"base": "<copy the base exactly>",

  ``variants'': {

      ``text'': ``'',

      ``not applicable'': false,

}}

Few-shot exemplars (follow style strictly):

-Base: <example statement> -Variant: <example variant>

Base statement: <base statement>

\end{lstlisting}

\begin{lstlisting}[caption={Prompt for it-clefts variants generation},breaklines=true,xleftmargin=0pt,basicstyle=\ttfamily\footnotesize]

You are a controlled text rewriter. Your only job is to transform the base statement into an It-cleft construction: It is/was [FOCUS] that [CLAUSE]. Generate in English only.

Task: Convert the base statement into an It-cleft variant.

 Hard constraints:
 (1) Use canonical It-cleft form: It is/was [FOCUS] that [CLAUSE]. Match the copula tense to the base.
 (2) [FOCUS] must be a contiguous verbatim span from the base. Allowed: NP (subject/object) focus, PP (time/place) focus, or adverbial. Not allowed: VP focus, paraphrase, removing content, insertion of new words.
 (3) [FOCUS] priority: object NP > PP > subject NP.
 (4) Keep the original article/none: a/an stays a/an; bare plurals/mass stay bare; definites remain definite.
 (5) Keep original PPs and word order inside the [CLAUSE] whenever possible.
 (6) Keep all named entities, numerals, negation, modals, quantifier scope, and PP complements unchanged.
 (7) Preserve truth conditions.
 (8) If the sentence is not suitable for an It-cleft, set ``not applicable''=true.

 Output format (SINGLE JSON only, no extra text):

 {"base": "<copy the base exactly>",

  ``variants'': {

      ``text'': ``'',

      ``not applicable'': false,

}}
  
Few-shot exemplars (follow style strictly):

-Base: <example statement> -Variant: <example variant>

Base statement: <base statement>


\end{lstlisting}

\begin{lstlisting}[caption={Prompt for wh-clefts variants generation:},breaklines=true,xleftmargin=0pt,basicstyle=\ttfamily\footnotesize]

You are a controlled text rewriter. Your only job is to transform the base statement into a Wh-cleft (pseudo-cleft) construction: What/Who/Where/When + [Clause with a gap] + is/are/was/were + [FOCUS]. Generate in English only.

Task: Convert the base statement into a Wh-cleft variant.

 Hard constraints:
 (1) Use canonical Wh-cleft form: What/Who/Where/When + [Clause with a gap] + is/are/was/were + [FOCUS]. Match the tense to the base.
 (2) WH choice: 'what' for things or VP-gaps (default), 'who' ONLY for people, 'where' for places, 'when' for times.
 (3) [FOCUS] must be a contiguous verbatim span from the base. Prefer NP or PP. VP allowed only if it is a contiguous phrase copied verbatim.
 (4) [FOCUS] preference order: object NP/PP > adjunct PP (time/place) > subject NP > VP.
 (5) Keep the original article/none: a/an stays a/an; bare plurals/mass stay bare; definites remain definite.
 (6) Keep all named entities, numerals, negation, modals, quantifier scope, and PP complements unchanged.
 (7) Preserve truth conditions.
 (8) If a well-formed Wh-cleft cannot be produced without paraphrasing or content change, set ``not applicable'' = true.

 Output format (SINGLE JSON only, no extra text):

 {"base": "<copy the base exactly>",

  ``variants'': {

      ``text'': ``'',

      ``not applicable'': false,

}}

Few-shot exemplars (follow style strictly):

-Base: <example statement> -Variant: <example variant>

Base statement: <base statement>


\end{lstlisting}

\begin{lstlisting}[caption={Prompt for SVC variants generation},breaklines=true,xleftmargin=0pt,basicstyle=\ttfamily\footnotesize]

You are a controlled text rewriter. Your only job is to transform the base statement into a Support Verb Construction (SVC): [SUPPORT VERB] + [DEVERBAL NOUN] (+ minimal required preposition) (+ original complements). Generate in English only.

Task: Convert the base statement into a Support Verb Constructions (SVC).

Hard constraints:
 (1) Make only the SVC substitution: [VERB] -> [SUPPORT VERB] + [DEVERBAL NOUN] (+ minimal required preposition) (+ original complements). Do not remove content. Do not make other paraphrasing.
 (2) If the verb in the base statement is modified by an adverb, project it as an adjective inside the SVC span if possible.
 (3) Keep all named entities, numerals, negation, modals, and quantifier scope unchanged.
 (4) Preserve complements by mapping them to the nominal head in a natural way; do not drop or invent content.
 (5) Aside from the SVC span and any required preposition, keep the rest of the wording identical. Preserve truth conditions.
 (6) If no SVC exists for the predicate, or if the base is already an SVC, set ``not applicable''=true.

 Output format (SINGLE JSON only, no extra text):

 {"base": "<copy the base exactly>",

  ``variants'': {

      ``text'': ``'',

      ``not applicable'': false,

}}

Few-shot exemplars (follow style strictly):

-Base: <example statement> -Variant: <example variant>

Base statement: <base statement>
\end{lstlisting}

\subsection{Responses Generation Prompt}
\label{sec:responses-prompt}

\begin{lstlisting}[caption={Prompt for it-clefts variants generation},breaklines=true,xleftmargin=0pt,basicstyle=\ttfamily\footnotesize]
You are a voter being asked for opinions. Your only job is to rate policy statements on a 1-7 Likert scale.

To what extent do you agree or disagree with the statement below? Please rate the statement using a 1-7 mapping score. Mapping: 1=Strongly disagree, 2=Disagree, 3=Slightly disagree, 4=Neutral, 5=Slightly agree, 6=Agree, 7=Strongly agree. Output one digit only.

Statement: <statement>

Score: 

\end{lstlisting}

\section{Evaluation Formulas}
\subsection{$W_1$ and Normalized Restoration Score Formulas}
\label{sec:w1-restoration-formulas}
For two 7-way distributions, $p$ and $q$, the distance can be computed in closed form via cumulative sums:
\begin{equation}
    CDF_p(d) = \sum_{i \leq d} p_i
\end{equation}
\begin{equation}
    W_1(p, q) = \sum_{d=1}^{7} |CDF_p(d) - CDF_q(d)|
\end{equation}
CDF here denotes the (discrete) cumulative distribution function over the ordered Likert categories.

\begin{table*}[t]
  \centering
  \begin{tabular}{ccccccc}
    \toprule
    Rewrite A & Rewrite B & Relation & r(G-4B) & r(12B) & r(Q-4B) & r(14B) \\ \midrule
    negation & opposite & invariance & \textbf{32.6\%} & 8.8\% & \textbf{25.4\%} & 6.5\% \\
    negation & active/passive & inversion & \textbf{28.3\%} & \textbf{20.8\%} & \textbf{25.4\%} & \textbf{19.5\%} \\
    negation & it-clefts & inversion & \textbf{30.5\%} & \textbf{29.5\%} & \textbf{22.5\%} & 13.5\% \\
    negation & wh-clefts & inversion & \textbf{35.2\%} & \textbf{33.0\%} & \textbf{48.3\%} & \textbf{28.6\%} \\
    negation & SVC & inversion & \textbf{28.4\%} & 10.0\% & 21.7\% & 8.3\% \\
    opposite & active/passive & inversion & \textbf{32.2\%} & \textbf{18.3\%} & \textbf{27.8\%} & \textbf{15.4\%} \\
    opposite & it-clefts & inversion & \textbf{33.3\%} & \textbf{18.7\%} & \textbf{27.5\%} & 9.4\% \\
    opposite & wh-clefts & inversion & \textbf{37.3\%} & \textbf{26.6\%} & \textbf{34.4\%} & \textbf{17.9\%} \\
    opposite & SVC & inversion & \textbf{37.3\%} & 22.2\% & \textbf{24.0\%} & 19.4\% \\
    it-clefts & active/passive & invariance & \textbf{15.2\%} & \textbf{1.9\%} & 8.9\% & \textbf{3.1\%} \\
    it-clefts & SVC & invariance & \textbf{16.1\%} & 0.0\% & \textbf{16.7\%} & \textbf{2.9\%} \\
    wh-clefts & active/passive & invariance & 12.6\% & 0.9\% & 7.8\% & 1.5\% \\
    wh-clefts & SVC & invariance & \textbf{14.0\%} & 0.0\% & \textbf{6.8\%} & \textbf{2.5\%} \\
    active/passive & SVC & invariance & \textbf{15.5\%} & 0.0\% & \textbf{8.3\%} & 0.0\% \\ \bottomrule
  \end{tabular}
  \caption{Results of combinations flip rate. Relation stands for expected relation of the statements pair. The bold font indicates the flip rate is higher than the individuals.}
  \label{tab-4}
\end{table*}

Let $p_c = z_D(x^{(c)})$ be the clean distribution, $p_k = z_D(x^{(k)})$ be the corrupt distribution, and $p_{int}$ be the distribution under an intervention (patch or ablation). The experiments define baseline distance as:
\begin{equation}
    d_0 = W_1(p_c, p_k)
\end{equation}
and define post-intervention distance as:
\begin{equation}
    d_{int} = W_1(p_c, p_{int})
\end{equation}
Then the restoration score is:
\begin{equation}
    R = 1 - \frac{d_{int}}{d_0}
\end{equation}

\subsection{Variance Decomposition Formulas}
\label{sec:VD-formulas}
Let $s$ denotes base statements, $v$ denotes variants, $r$ denotes repeated samples, and $E$ denotes the expectation. For each $(s, v, r)$ the model generates a 1-7 Likert score $y_{s,v,r}$, define
\begin{equation}
  \bar{y}_{s,v} = E_r[y_{s,v,r}]
\end{equation}
and
\begin{equation}
  \bar{y}_s = E_v[y_{s,v}]
\end{equation}
Purpose Sensitivity (PS) is calculated by using the formula:
\begin{equation}
    PS = \frac{1}{|S_v|-1}\sum_{s\in S_v}(\bar{y}_s - \frac{1}{|S_v|}\sum_{s'\in S_v}\bar{y}_{s'})^2
\end{equation}
Articulation Sensitivity (AS) is calculated by applying the formula:
\begin{equation}
    AS = \frac{1}{|S_v|}\sum_{s\in S_v}\frac{(\bar{y}_{s,orig} - \bar{y}_{s,v})^2}{2}
\end{equation}
Model Uncertainty (MU) is calculated by using the formula:
\begin{equation}
  MU = E_{s,v}[Var_r(y_{s,v,r})]
\end{equation}

\section{Combinations of Linguistic Rewrites}
\label{sec:combinations}
Naturally occurring statements may contain multiple simultaneous linguistic phenomena (e.g., negation + opposite), and robustness failures might therefore arise not only from individual edits but also from their interactions. We therefore also test whether the stance instability of combinations is higher than that of the individual linguistic rewrites.

For the six linguistic rewrites, we create all pairwise combinations wherever the two transformations can be applied without creating a structurally ill-formed sentence. The final set contains 14 valid pairwise combinations. Combined variants were generated using the same few-shot prompting strategy as in Section~\ref{sec:data} and evaluated with the same methods as the individual edits.

The flip results of the models for linguistic rewrites combinations are shown in Table~\ref{tab-4}. The combinations' flip rates are generally higher than that of the individual linguistic rewrites, indicating that stance instability of combinations could be higher than that of the single rewrites.

\section{Activation Patching Results of Qwen3}
\label{sec:qwen3-patching}
\begin{figure*}[t]
  \centering
  \includegraphics[width=0.45\linewidth]{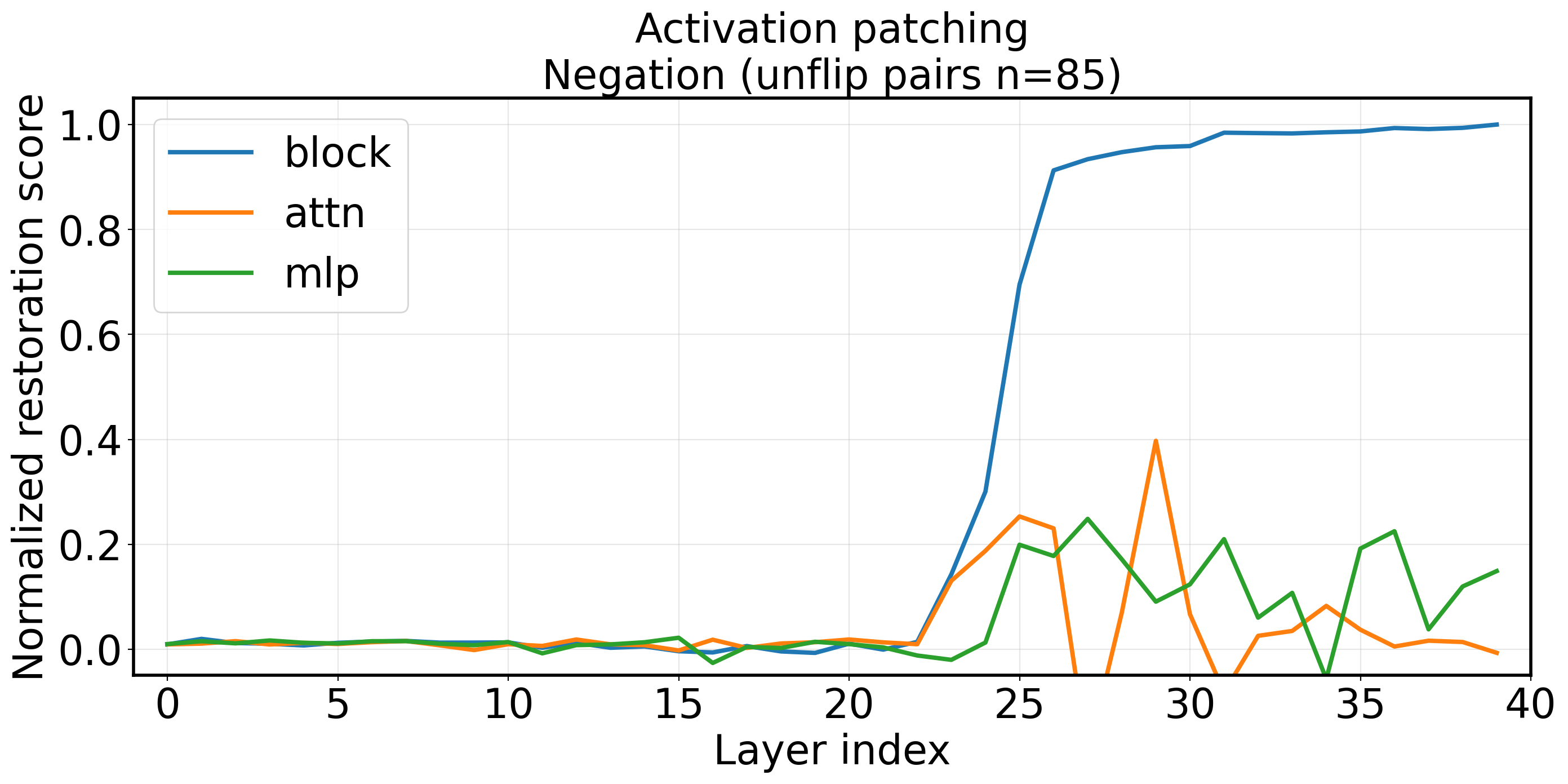}
  \includegraphics[width=0.45\linewidth]{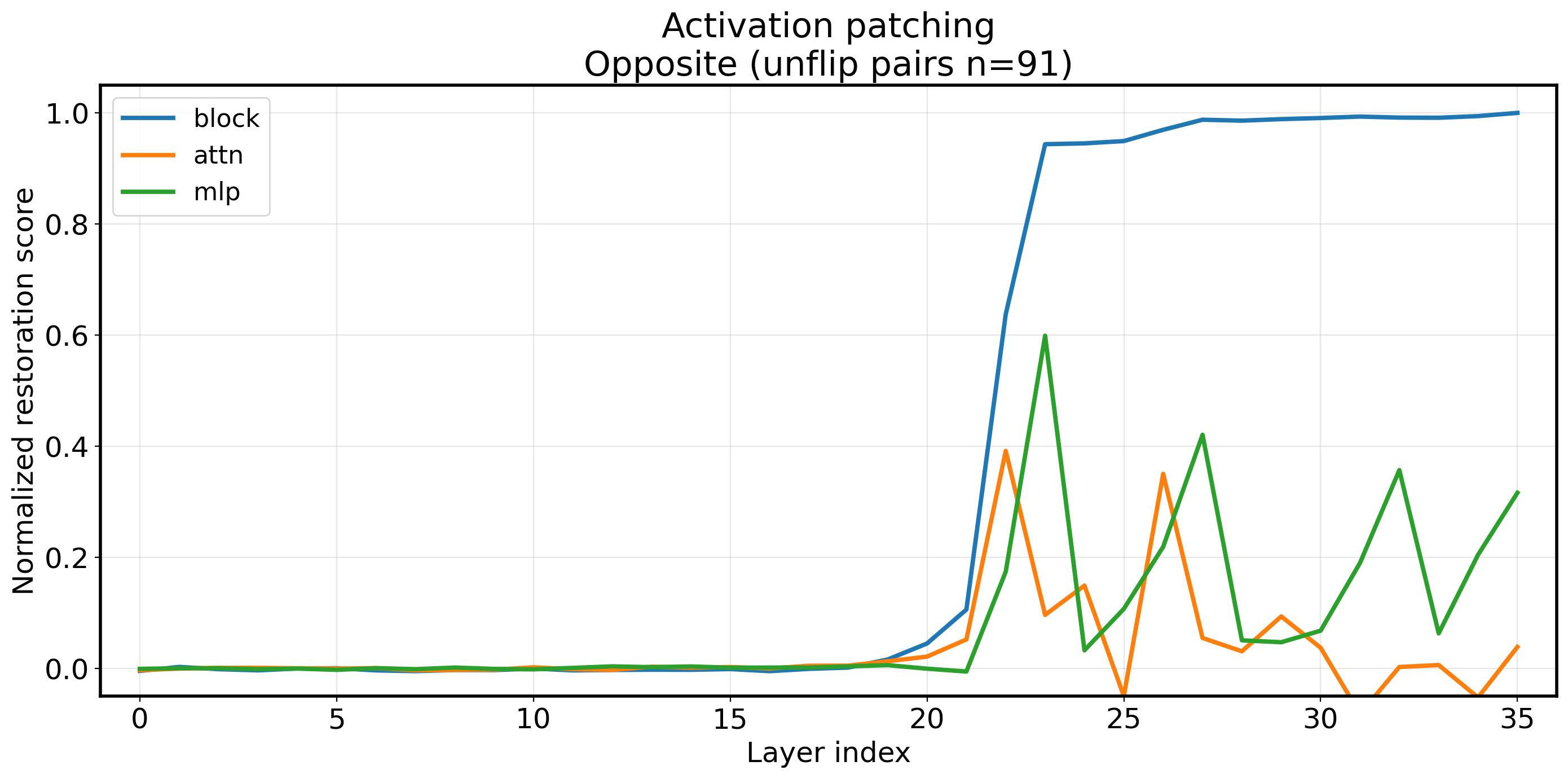}
  \includegraphics[width=0.45\linewidth]{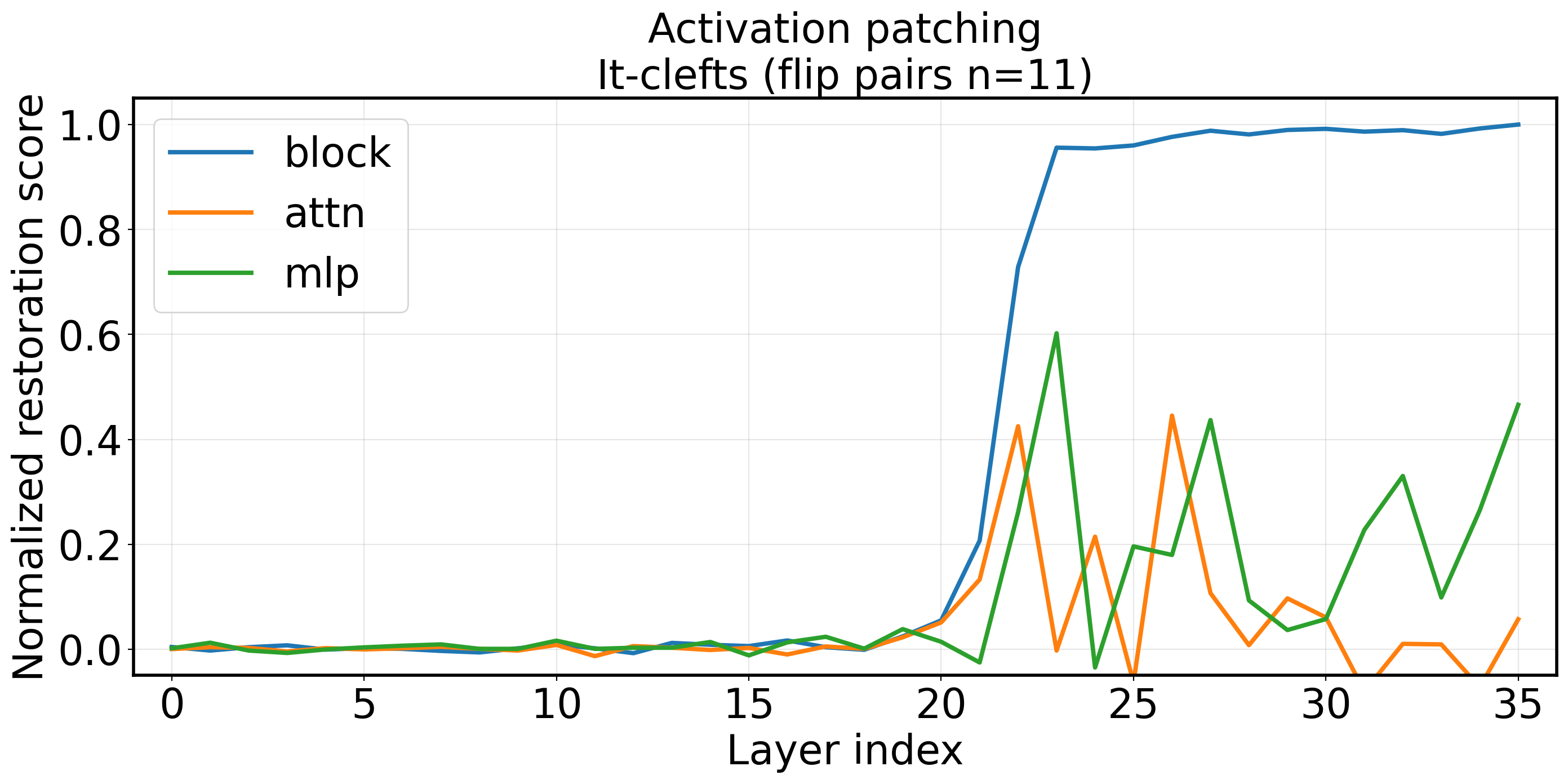}
  \includegraphics[width=0.45\linewidth]{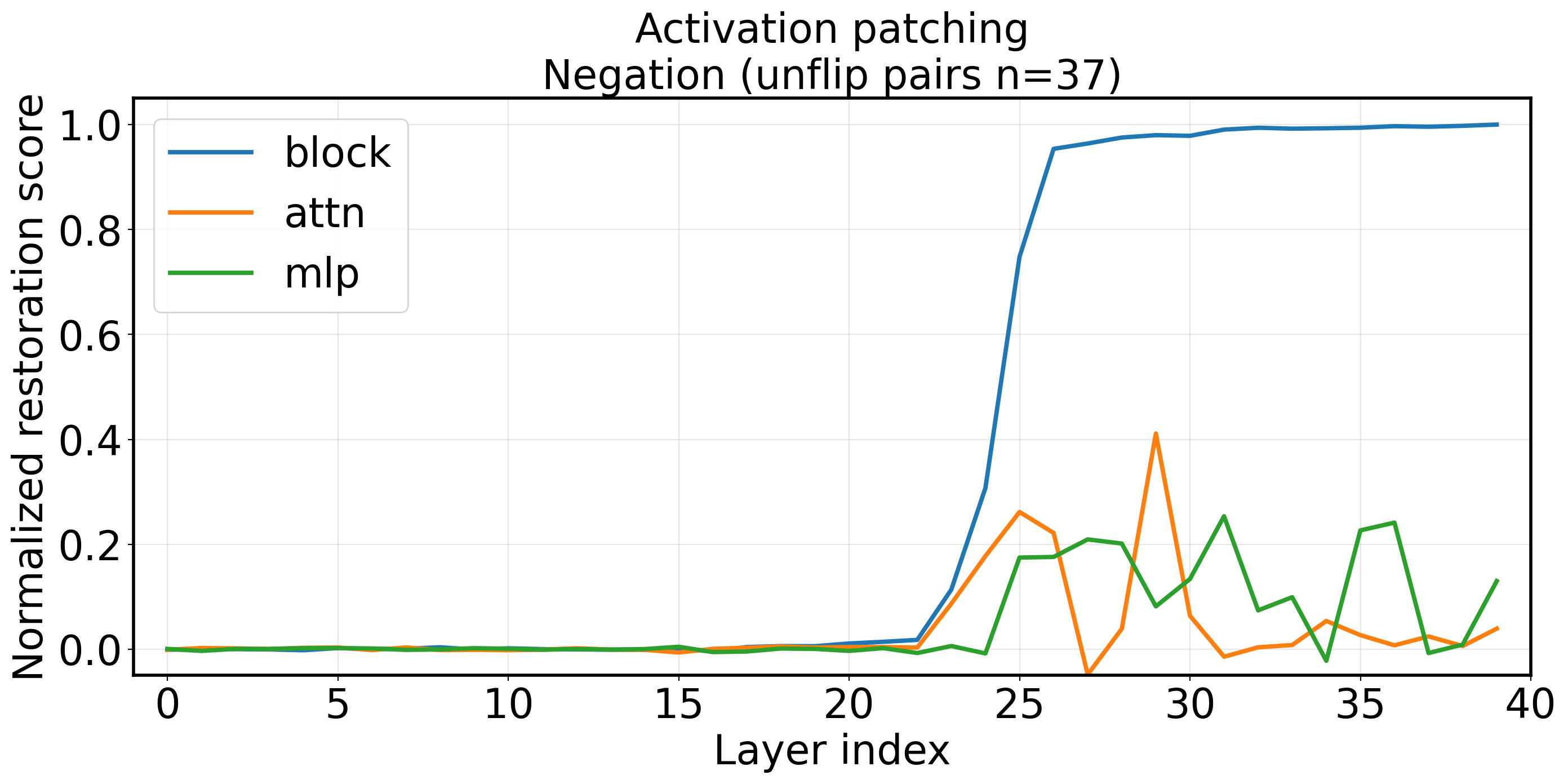}
  \includegraphics[width=0.45\linewidth]{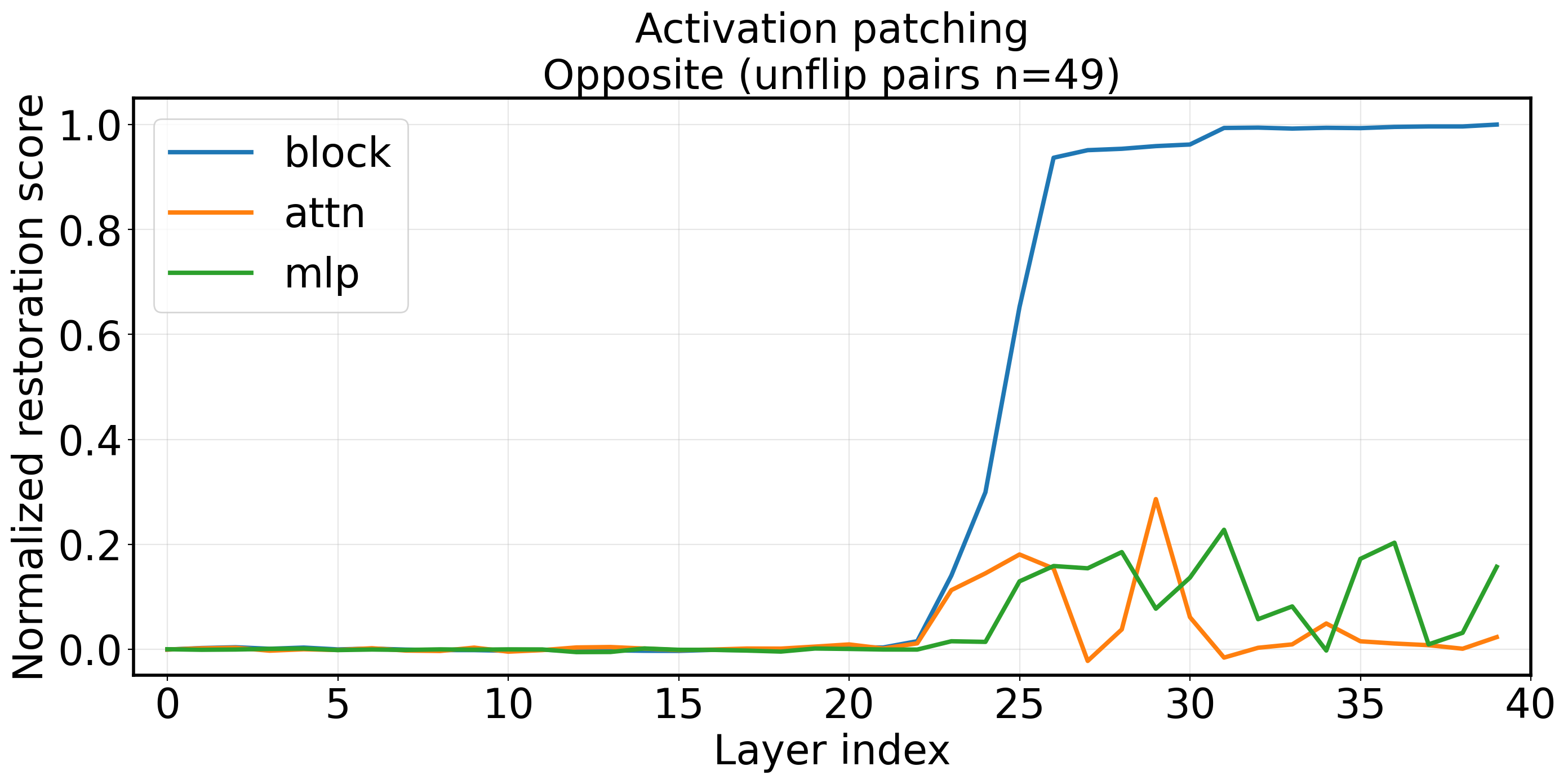}
  \caption{Activation patching results of the Qwen3-4B and Qwen3-14B model. The figures in the first row and the left one on the second row are the results of Qwen3-4B, other two figures are the outcome of Qwen3-14B.}
  \label{fig:qwen-patching-results}
\end{figure*}

Please see Figure~\ref{fig:qwen-patching-results}. The five subfigures show the activation patching results of the Qwen3-4B and Qwen3-14B model.

\end{document}